\documentclass[letterpaper, 10 pt, conference]{ieeeconf}  
\IEEEoverridecommandlockouts
\usepackage{bm}
\usepackage{url}
\usepackage{tikz}
\usepackage{xargs}
\usepackage{ifthen}
\usepackage{siunitx}
\usepackage{contour}
\usepackage{amsmath}
\usepackage{amssymb}
\usepackage{amsfonts}
\usepackage{booktabs}
\usepackage{graphicx}
\usepackage{multicol}
\usepackage{multirow}
\usepackage{pgfplots} 
\usepackage{verbatim}
\usepackage{textcomp}
\usepackage{xinttools}
\usepackage[nolist]{acronym}
\usepackage[dvipsnames]{xcolor}
\usepackage[hidelinks]{hyperref}
\graphicspath{{figures/}}
\pgfplotsset{compat=1.18}
\usetikzlibrary{calc}
\usetikzlibrary{bending}
\usetikzlibrary{shadows.blur}
\usetikzlibrary{positioning}
\usetikzlibrary{arrows.meta}
\usetikzlibrary {shapes.geometric}
\title{\LARGE \bf Self-supervised Domain Adaptation for Visual 3D Pose Estimation of Nano-drone Racing Gates by Enforcing Geometric Consistency}

\author{Nicholas Carlotti$^{1*}$, Michele Antonazzi$^{2*}$, Elia Cereda$^{1}$, Mirko Nava$^{1}$,\\Nicola Basilico$^2$, Daniele Palossi$^{13}$, and Alessandro Giusti$^{1}$
\thanks{$^*$ Indicates equal contribution.}
\thanks{$^{1}$Nicholas Carlotti, Elia Cereda, Mirko Nava, Daniele Palossi, and Alessandro Giusti are with the Dalle Molle Institute for Artificial Intelligence (IDSIA), USI-SUPSI, Lugano, Switzerland {\tt\small nicholas.carlotti@supsi.ch}}%
\thanks{$^{2}$Michele Antonazzi and Nicola Basilico are with University of Milan, Italy {\tt\small<name>.<surname>@unimi.it}}%
\thanks{$^{3}$Daniele Palossi is also with the Integrated Systems Laboratory (IIS), ETHZ, Z\"urich, Switzerland.}%
\thanks{This work is supported by the Swiss National Science Foundation, grant number 213074.}
}
    
\begin{document}

\maketitle
\thispagestyle{empty}
\pagestyle{empty}

\definecolor{purple_pizzaz}{HTML}{FE4EDA}%
\definecolor{giants_orange}{HTML}{FE5A1D}%
\definecolor{wong_gray}{HTML}{888888}%
\definecolor{wong_black}{HTML}{333333}%
\definecolor{wong_gold}{HTML}{E69F00}%
\definecolor{wong_cyan}{HTML}{56B4E9}%
\definecolor{wong_green}{HTML}{009E73}%
\definecolor{wong_yellow}{HTML}{F0E442}%
\definecolor{wong_blue}{HTML}{0072B2}%
\definecolor{wong_red}{HTML}{D55E00}%
\definecolor{wong_pink}{HTML}{CC79A7}%
\definecolor{wong_magenta}{HTML}{CA1963}%
\newcommand{\elia}{{\color{wong_cyan!80!black}\bf @Elia}}
\newcommand{\notes}[1]{\textcolor{black}{#1}}

\newcommand{\gatep}{\ensuremath{\mathcal{P}_\text{G}}}
\newcommand{\gatephat}{\ensuremath{\mathcal{\hat{P}}_\text{G}}}
\newcommand{\droneo}{\ensuremath{\mathcal{\hat{O}}}}
\newcommand{\droneogt}{\ensuremath{\mathcal{O}}}
\newcommand{\image}{\ensuremath{\mathcal{I}}}
\newcommand{\robotpose}{\ensuremath{\mathcal{P}_\text{R}}}
\newcommand{\robotposehat}{\ensuremath{\mathcal{\hat{P}}_\text{R}}}

\newcommand{\datas}{\ensuremath{\mathcal{D}^\text{train}_\text{sim}}}
\newcommand{\datar}{\ensuremath{\mathcal{D}^\text{train}_\text{real}}}
\newcommand{\datasval}{\ensuremath{\mathcal{D}^\text{val}_\text{sim}}}
\newcommand{\datastrain}{\ensuremath{\mathcal{D}^\text{train}_\text{sim}}}

\newcommand{\datarval}{\ensuremath{\mathcal{D}^\text{val}_\text{real}}}
\newcommand{\datartest}{\ensuremath{\mathcal{D}^\text{test}_\text{real}}}
\newcommand{\datartrain}{\ensuremath{\mathcal{D}^\text{train}_\text{real}}}

\newcommand{\model}[1][]{\ensuremath{m\def\temp{#1} \ifx\temp\empty\else (#1) \fi}}

\newcommand{\lossp}{\ensuremath{\mathcal{L}_\text{pose}}}
\newcommand{\losssc}{\ensuremath{\mathcal{L}_\text{sc}}}

\newcommand{\shorteq}{%
  \resizebox{2mm}{\height}{=}%
}

\DeclareSIUnit{\mac}{MAC}
\DeclareSIUnit{\nothing}{\relax}

\begin{abstract}
We consider the task of visually estimating the relative pose of a drone racing gate in front of a nano-quadrotor, using a convolutional neural network pre-trained on simulated data to regress the gate's pose.  Due to the sim-to-real gap, the pre-trained model underperforms in the real world and must be adapted to the target domain. We propose an unsupervised domain adaptation (UDA) approach using only real image sequences collected by the drone flying an arbitrary trajectory in front of a gate; sequences are annotated in a self-supervised fashion with the drone's odometry as measured by its onboard sensors. On this dataset, a state consistency loss enforces that two images acquired at different times yield pose predictions that are consistent with the drone's odometry. Results indicate that our approach outperforms other SoA UDA approaches, has a low mean absolute error in position ($\bm{x}$=26, $\bm{y}$=28, $\bm{z}$=10 cm) and orientation ($\bm{\psi}$=13$\bm{^{\circ}}$), an improvement of $\bm{40\%}$ in position and $\bm{37\%}$ in orientation over a baseline.
The approach's effectiveness is appreciable with as few as 10 minutes of real-world flight data and yields models with an inference time of 30.4ms (33 fps) when deployed aboard the Crazyflie 2.1 Brushless nano-drone.
\end{abstract}

\section{Introduction}\label{sec:intro}
Training deep learning models for perception tasks requires large datasets of annotated data.
Obtaining real world data at scale is difficult as the collection process is often expensive and time-consuming.
Conversely, simulators allow for easy collection of large amounts of training data; however, differences with the real world in appearance and dynamics pose a challenge, the so-called sim-to-real gap~\cite{salvato2021crossing}.
To bridge the gap between simulated and real-world data, models are be adapted with domain transfer approaches~\cite{schwonberg2023survey,li2024comprehensive}, with \textit{domain} indicating a specific data distribution, e.g., simulated and real-world environments.

\begin{figure}[!t]%
    \centering%
    \input{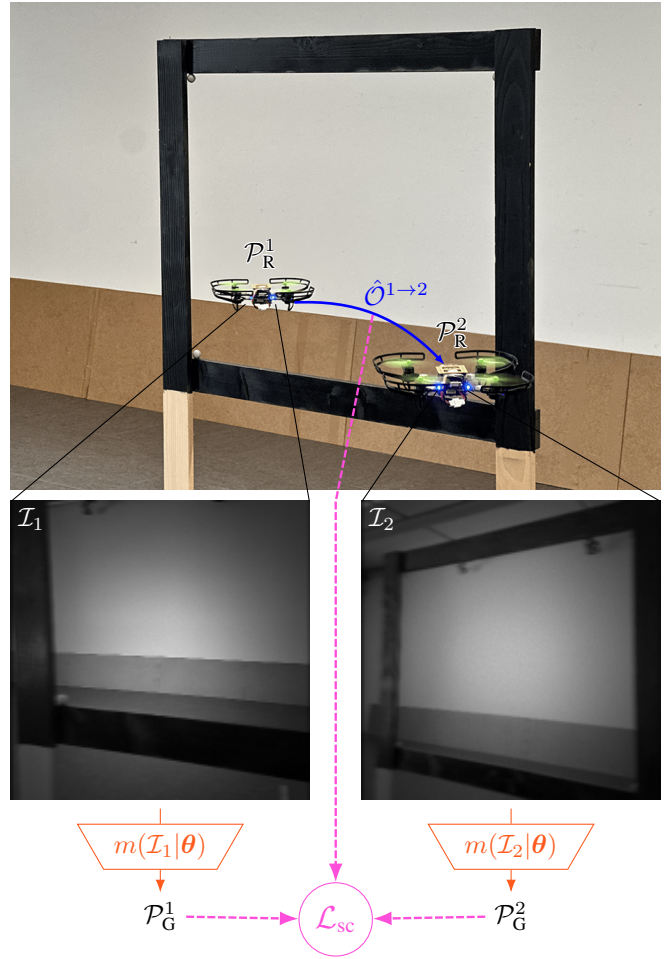}%
    \caption{We estimate the pose ($x$, $y$, $z$, $\psi$) of drone racing gates by transferring a model trained in simulation to a real-world environment and deploying it aboard the Bitcraze Crazyflie 2.1 Brushless nano-UAV.}%
    \label{fig:intro}%
\end{figure}

We focus on a drone-racing scenario, in which competing drones have to pass through gates with a characteristic appearance in a precise order to win a race (see Figure~\ref{fig:intro}).
In this scenario, we specifically consider the task of monocular visual pose estimation of drone-racing gates in a nano-drone~\cite{lamberti2024droneracingsimtoreal}, starting from a model pre-trained in simulation, and adapting it to the real world.
Given a single image taken by a drone's onboard camera, our model estimates the 3D pose of a gate relative to the drone.
We adopt the Bitcraze Crazyflie 2.1 Brushless\footnote{https://www.bitcraze.io/products/crazyflie-2-1-brushless} nano-drone platform, which mounts a grayscale, low-resolution camera with a limited dynamic range:
Images are noisy, affected by motion blur and vignetting, and, crucially, significantly differ from those acquired in simulation, which lack these artifacts.
A naive adaptation solution is fine-tuning the model on target domain data labeled with ground truth gate poses~\cite{scarciglia2025gatedetection}; however, this approach requires a motion capture system to collect pose data.
Other solutions consist in zero-shot transfer of the task~\cite{ lamberti2024droneracingsimtoreal,pham2022pencilnet}, but are ineffective when the domain shift is too large.
Self-supervised approaches from the computer vision literature~\cite{nguyen22self, yoon22lassification} represent a valid alternative, yet they fail to leverage the full potential of robotic platforms and their sensory apparatus.
An unexplored approach for drone racing is robot self-supervision~\cite{nava2019learning, sathyamoorthy2022terrapn, seo2023scate}, where training data is collected and autonomously labeled by the robots themselves without the need for human supervision.

We propose an approach for sim-to-real domain transfer based on fine-tuning on autonomously-collected target domain data:
A nano-drone flying random trajectories in front of a gate acquires images and odometry with its onboard sensors.
We derive a supervision signal from the odometry data by enforcing a geometric state consistency loss~\cite{nava2021state} between pairs of poses predicted by the model:
Given two images of the gate taken from different locations, our loss imposes that the two predicted relative gate poses must be compatible with the drone's measured movement between the two locations (see Figure~\ref{fig:intro}).
Compared to general domain adaptation approaches, our work is designed for robotic platforms and uses the robot's sensors \notes{and odometry estimation} as the basis of the approach.
This renders the approach effective at domain transfer of vision tasks, such as estimating the pose of a drone racing gate.
Moreover, the real-world data required by our approach is cheaply obtained using a single drone equipped with a camera \notes{and the onboard odometry estimation}.

After discussing related works in domain adaptation and self-supervised robot learning in Section~\ref{sec:rw}, we present our main \textbf{contribution} in Section~\ref{sec:method}: a novel application of the state consistency loss for the domain transfer of a 3D gate pose estimation task; the loss is combined with a self-supervised data collection pipeline making the approach practical and easily applicable in any target domain.
In Section~\ref{sec:setup}, we describe the robotic platform, simulated and real-world domains, data collection pipeline, and evaluation metrics.
Results, illustrated in Section~\ref{sec:results}, show that our approach effectively transfers perception capabilities from simulation to the real-world, even with as few as 10 minutes of real-world flight data.
Specifically, our method achieves a position error of$10$ cm and an orientation error of $13\bm{^{\circ}}$, improving over SoA UDA baselines by 40\% and 37\%, respectively.
Furthermore, we assess the effectiveness of our approach on a small neural network tailored for running at 33 FPS onboard a nano-drone platform. 
Conclusions are drawn in Section~\ref{sec:conclusion} along with future work.

\section{Related Work}\label{sec:rw}

\subsection{Gate Detection for Drone Racing}

Accurate gate detection is fundamental to enable fast navigation in drone racing competitions~\cite{lamberti2024droneracingsimtoreal,foehn2022alphapilot}. 
%
%
%
The straightforward approach to train neural networks for gate detection is to directly use data collected with a motion capture system~\cite{scarciglia2025gatedetection}.
However, these systems are expensive and unsuitable for use across multiple environments.
As such, alternative approaches learn from simulated data while mitigating the sim-to-real gap with strong image augmentations~\cite{lamberti2024droneracingsimtoreal}, or by randomizing the visual aspect of many simulated environments~\cite{loquercio2020deepdroneracing}.
Another way to align the visual appearance of simulated and real images is to apply a pencil filter that highlights edges and disregards flat image regions~\cite{pham2022pencilnet}.
Instead, we propose an approach for sim-to-real transfer using images from the target environment, which utilizes the drone's odometry as supervision for the alignment process. 

\subsection{Domain Generalization}
In domain generalization, training data is collected in multiple source domains to cover a large portion of the input space, an effective strategy to generalize models to unseen domains.
A prominent approach is to randomize the domain using simulators by altering the number and location of objects, modifying illumination conditions, and randomizing textures to match the complexity of real-world environments~\cite {tobin2018domainrando, loquercio2020deepdroneracing}.
Other approaches leverage data generation to synthesize randomized, realistic training data~\cite{rahman2019gandomaingeneralization,james2019gangrasping} or alter the input space to render the two domains indistinguishable with autoencoders~\cite{inoue2018transfer}, or a generative adversarial network~\cite{james2019sim}.
A simpler approach for making domains indistinguishable is to preprocess input images with a pencil filter to emphasize edges and remove distracting domain details~\cite{pham2022pencilnet}.

Instead of generalizing to any domain, our approach acquires fine-tuning data in the target domain directly and adapts to it by optimizing the state consistency loss.

\subsection{Domain Adaptation}
When data can be collected in the target domain, \emph{domain adaptation} approaches fine-tune a model to capture the specific characteristics of the domain; the simple solution is to use high-quality labeled data, often involving laborious hand-labeling~\cite{antonazzi2024roboticvision, zimmeramn2023longtermlocalization} or ad-hoc sensing apparatus \cite{scarciglia2025gatedetection}.
%

Performing adaptation in \emph{unsupervised} fashion, i.e, without ground-truth annotations for the target domain, is an attractive alternative~\cite{li2024comprehensive}.
Two common solutions are to reduce the domain shift or learn a latent space in the target domain with self-supervised learning.
Approaches reduce the domain shift at the input level, rendering the source and target input data indistinguishable by performing style transfer between domains~\cite{zhu2017cyclegan} and further aligning image features in a latent space~\cite{hoffman2018cycada, chen2019crdoco}.
Reducing the domain shift at feature level is done with adversarial learning, by confusing a discriminator network~\cite{tzeng2019adversarialdomainadaptation, zhang2020jointadversarialoutputadaptation}, or by minimizing a notion of domain divergence~\cite{jingdong2018distributionalignement, fua2019sharingweights}, such as the Maximum Mean Discrepancy (MMD)~\cite{mmd}. 
%
%
%
Self-supervised domain adaptation without target domain labels is done with teacher-student training of a model pre-trained on source data, to minimize the model entropy on the target domain~\cite{kim2021domain}.
Approaches update self-labels at each training iteration to mitigate noisy predictions~\cite{kothandaraman2021ss} or update only labels for high-uncertainty samples~\cite{chen2022self,chu2022denoised}, weight the loss by the model uncertainty under different input augmentations~\cite{yan2021augmented}, or minimize a triplet loss~\cite{wang2022cross}.

Compared to other unsupervised approaches, we do not use source domain data for model fine-tuning; instead, we use unlabeled target domain data to optimize a consistency loss specifically formulated for odometry-capable robots.
%

\subsection{Self-supervised Robot Learning}

%
To lessen the need for labeled training data, self-supervised approaches use trivial learning objectives, whose labels are generated by transforming the input data, leveraging the many sensors equipped on a robot platform.
Typical examples include the use of proximity sensors and odometry to self-supervise the training of an obstacle detection model~\cite{nava2019learning}, or dense point clouds for self-supervised traversability estimation~\cite{seo2023scate}.
Recent self-supervised approaches leverage a pretext task, whose objective is not to directly solve the task of interest but, rather, to learn similar pattern recognition skills as those needed to solve the task of interest:
approaches estimate, given an input image, the sound produced by a drone~\cite{nava2022learning}, or the state of the LEDs fitted on a robot~\cite{carlotti2024learning}; others solve masked autoencoding tasks on images~\cite{radosavovic2023real}, or use two sensory streams such as vision and language~\cite{karamcheti2023language}, or vision and haptics~\cite{liu2024masked}.

Our approach leverages data collected autonomously by the drone to directly supervise a deep learning model on the transfer from simulation to reality of a drone racing gate detector.

\section{Methodology}\label{sec:method}
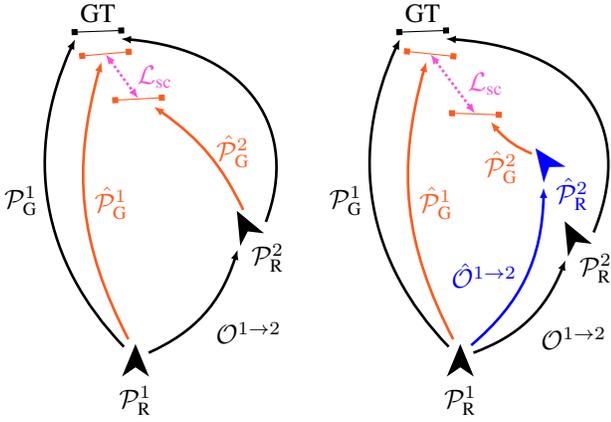
\begin{figure}[tp]%
    \centering%
    \hfill\begin{tikzpicture}
\newcommand{\gate}[4]{%
\coordinate (#1) at #2;
    \draw[{Square[scale=0.7]}-{Square[scale=0.7]}, #3] ($ (#1) - (0.35,0.0) $) -- node [above] {#4} ($ (#1) + (0.35,0.0) $) {};
}
\begin{scope}[shift={(-0.5\linewidth, -4.0)}]
\node [scale=0.6, fill=black, dart, rotate=90, dart tail angle=120] (robot1) at (0.0,0.0) {};
\node [below=2mm of robot1.text, anchor=north] {$\robotpose^{1}$}; 
\node [scale=0.6, fill=black, dart, rotate=118, dart tail angle=120] (robot2) at (1.46,1.8) {};
\node [below right=1mm and 3mm of robot2.base, anchor=north] {$\robotpose^{2}$}; 
\gate{gate}{(-0.5,4.4)}{black, rotate=2}{GT} 
\gate{gatepred1}{(-0.40,4.1)}{giants_orange, rotate=5}{} 
\gate{gatepred2}{(0.04,3.5)}{giants_orange, rotate=3}{} 
\draw [black, -{Latex[length=4pt,width=3pt]}, line width=1.0pt] ($ (robot1.left side) + (-0.06, 0.18) $) to [bend left=35] node [left, midway] {$\gatep^1$} ($ (gate) - (0.3, 0.1) $); 
\draw [black, -{Latex[length=4pt,width=3pt]}, line width=1.0pt] ($ (robot2.right side) + (0.2, 0.0) $) .. controls (2.2,3.1) and (1.5,4.2) .. ($ (gate) + (0.3, -0.1) $); 
\draw [giants_orange, -{Latex[length=4pt,width=3pt]}, line width=1.0pt] ($ (robot1.left side) + (0.0, 0.28) $) to [bend left=22] node [right, midway] {$\gatephat^1$} ($ (gatepred1) - (0.05, 0.1) $); 
\draw [giants_orange, -{Latex[length=4pt,width=3pt]}, line width=1.0pt] ($ (robot2.right side) + (-0.1, 0.16) $) to [bend right=15] node [right, midway] {$\gatephat^2$} ($ (gatepred2) + (0.2, -0.1) $); 
\draw [black, -{Latex[length=4pt,width=3pt]}, line width=1.0pt] ($ (robot1.right side) + (0.08, 0.1) $) to [bend right=25] node [below right, midway] {$\droneogt^{1\rightarrow 2}$} ($ (robot2.left tail) - (0.03, 0.03) $); 
\draw [purple_pizzaz, {Latex[length=4pt,width=3pt]}-{Latex[length=4pt,width=3pt]}, dashed, dash pattern=on 1pt off 1pt, line width=1.0pt] ($ (gatepred1) + (0.0, 0.0) $) to [bend right=0] node [right, midway, xshift=2pt] {\losssc} ($ (gatepred2) + (0.0, 0.0) $); 
\end{scope}
\begin{scope}[shift={(0, -4.0)}]
\node [scale=0.6, fill=black, dart, rotate=90, dart tail angle=120] (robot1) at (0,0) {};
\node [below=2mm of robot1.text, anchor=north] {$\robotpose^{1}$}; 
\node [scale=0.6, fill=black, dart, rotate=118, dart tail angle=120] (robot2) at (1.5,1.66) {};
\node [below right=1mm and 3mm of robot2.base, anchor=north] {$\robotpose^{2}$}; 
\node [scale=0.6, fill=blue, dart, rotate=126, dart tail angle=120] (robot2pred) at (1.14,2.65) {};
\node [below right=0.8mm and 3.5mm of robot2pred.base, anchor=north, blue] {$\robotposehat^{2}$}; 
\gate{gate}{(-0.5,4.4)}{black, rotate=2}{GT} 
\gate{gatepred1}{(-0.40,4.1)}{giants_orange, rotate=-6}{} 
\gate{gatepred2}{(0.2,3.3)}{giants_orange, rotate=-2}{} 
\draw [black, -{Latex[length=4pt,width=3pt]}, line width=1.0pt] ($ (robot1.left side) + (-0.06, 0.18) $) to [bend left=35] node [left, midway] {$\gatep^1$} ($ (gate) - (0.3, 0.1) $); 
\draw [black, -{Latex[length=4pt,width=3pt]}, line width=1.0pt] ($ (robot2.right side) + (0.2, 0.0) $) .. controls (2.3,3.0) and (1.8,4.2) .. ($ (gate) + (0.3, -0.1) $); 
\draw [giants_orange, -{Latex[length=4pt,width=3pt]}, line width=1.0pt] ($ (robot1.left side) + (0.0, 0.28) $) to [bend left=22] node [right, midway] {$\gatephat^1$} ($ (gatepred1) - (0.05, 0.1) $); 
\draw [giants_orange, -{Latex[length=4pt,width=3pt]}, line width=1.0pt] ($ (robot2pred.left side) + (-0.1, 0.16) $) to [bend left=15] node [below, midway, shift=(0:-0.1)] {$\gatephat^2$} ($ (gatepred2) + (0.2, -0.1) $); 
\draw [black, -{Latex[length=4pt,width=3pt]}, line width=1.0pt] ($ (robot1.right side) + (0.08, 0.1) $) to [bend right=25] node [below right, midway] {$\droneogt^{1\rightarrow 2}$} ($ (robot2.left tail) - (0.03, 0.03) $); 
\draw [blue, -{Latex[length=4pt,width=3pt]}, line width=1.0pt] ($ (robot1.right side) + (0.05, 0.2) $) to [bend right=25] node [left, midway] {$\droneo^{1\rightarrow 2}$} ($ (robot2pred.left tail) - (0.0, 0.05) $); 
\draw [purple_pizzaz, {Latex[length=4pt,width=3pt]}-{Latex[length=4pt,width=3pt]}, dashed, dash pattern=on 1pt off 1pt, line width=1.0pt] ($ (gatepred1) + (0.0, 0.0) $) to [bend right=0] node [right, midway, xshift=2pt] {\losssc} ($ (gatepred2) + (0.0, 0.0) $); 
\end{scope}
\end{tikzpicture}
    \caption{Assuming perfect odometry (left), the state consistency loss forces the relative gate poses $\gatephat^1$ and $\gatephat^2$, predicted from the drone's poses at $\robotpose^{1}$ and $\robotpose^{2}$, to be coherent with the drone's relative odometry $\droneogt^{1\rightarrow 2}$ between the two poses.
    In our experiments, the measured odometry $\droneo^{1\rightarrow 2}$ is collected by the drone itself and is affected by drift and noise (right).}
    \label{fig:method}%
\end{figure}
We solve the problem of estimating the 6D pose of a drone racing gate relative to an observing drone $\gatep \in SE(3)$, given a single grayscale image \image.
The model $\model[\image|\bm{\theta}]$ parametrized by weights $\bm{\theta}$ takes a single image \image\! as input and produces the estimated gate pose \gatephat.
%
%
Initially, the model is trained on data coming from a simulator, allowing us to easily collect images labeled with the pose of the simulated gate \gatep.
In a second phase, we transfer the model trained in simulation to the real world by fine-tuning on data collected by a real drone flying in front of a gate, annotated with the drone's odometry while having no access to information about the gate and its pose. The drone odometry collected at time $t=1$ w.r.t. the world reference frame is denoted by $\droneo^{1\rightarrow w}$. 
The resulting simulated dataset is denoted with $\mathcal{D}_\text{sim}$ and the real-world one with $\mathcal{D}_\text{real}$.
The training procedure consists of a first phase where the model is trained on \datas, while in the second phase, we start from the model parameters minimizing \lossp\! on the simulated validation set \datasval\! and fine-tune them with \datar\! using only the \losssc\! loss.

When training on simulated data, we minimize the mean squared error (MSE) loss between \gatep\! and \gatephat:

\begin{equation}
    \lossp = \text{MSE}(\gatep, \gatephat)
\end{equation}
where the position components of the pose are directly passed to the MSE loss, while the orientation components are first converted to the 6D representation proposed in Zhou et al.~\cite{zhou2019continuity}, preserving the continuity of 3D rotations to aid the training process.

For real-world data, we enforce a State Consistency (SC) loss~\cite{nava2021state} between two model predictions, exploiting the known drone odometry:
Given a pair of images taken by the drone in a static scene from two different locations, the difference between the estimated relative gate's poses as perceived by the drone in the two locations must be equal to the drone's movement between the first and second locations (see Fig.~\ref{fig:method} for a general overview).
Formally, given two input images $\image_{1}, \image_{2}$, the gate poses $\gatep^1, \gatep^2$, and the drone odometries $\droneo^{1\rightarrow w}, \droneo^{2\rightarrow w}$, we first make the model predict $\gatephat^1, \gatephat^2$. Then, we warp $\gatephat^1$ to the frame of reference of the drone at the position in which $\image_{2}$ was taken
%
$
\gatephat^{1 \rightarrow 2} = \gatephat^1 \cdot \droneo^{2\rightarrow w} \cdot (\droneo^{1\rightarrow w})^{-1}
$,
%
where $\cdot$ denotes matrix multiplication and $\mathcal{M}^{-1}$ denotes the inverse of matrix $\mathcal{M}$.
The SC loss is then written as:
\begin{equation}
    \losssc = \text{MSE}(\gatephat^{1 \rightarrow 2}, \gatephat^2)
    \label{eq:stateconsistency}
\end{equation}
where orientation components of the poses are converted to the 6D representation proposed in Zhou et al.~\cite{zhou2019continuity}.

\section{Experimental Evaluation}\label{sec:setup}

\begin{figure}[tp]%
    \centering%
    \includegraphics[width=0.6\linewidth]{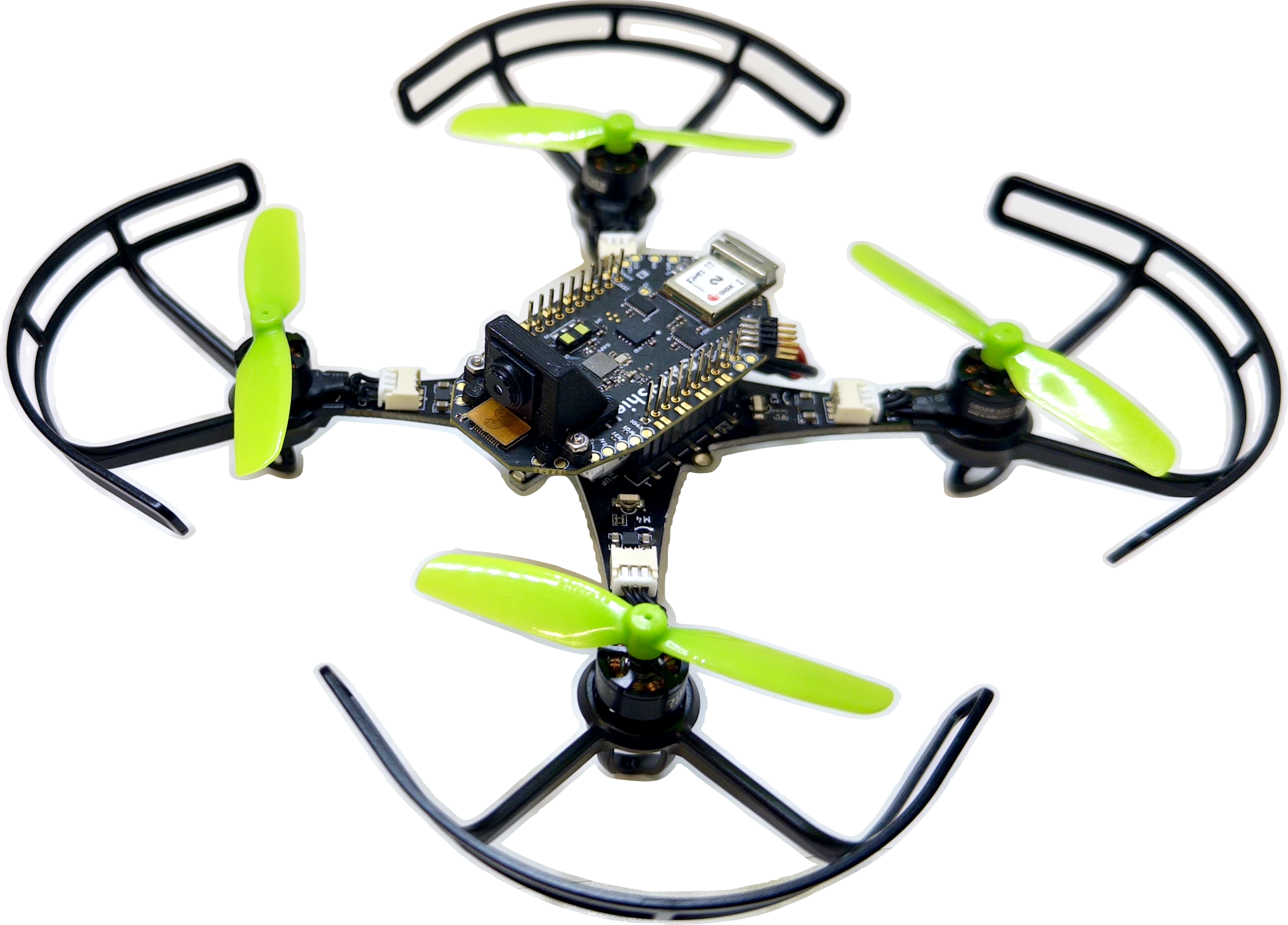}%
    \caption{The Crazyflie 2.1 Brushless nano-drone used in our experiments.}%
    \label{fig:cf}%
\end{figure}

\subsection{Robot Platform}\label{sec:robot-platform}

\begin{figure*}[thbp]
    \centering
    \includegraphics[width=0.165\linewidth]{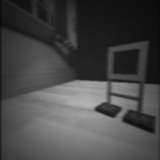}%
    \hfill
    \includegraphics[width=0.165\linewidth]{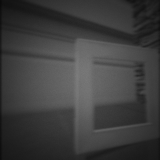}%
    \hfill
    \includegraphics[width=0.165\linewidth]{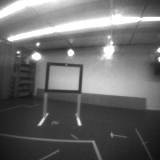}%
    \hfill
    \includegraphics[width=0.165\linewidth]{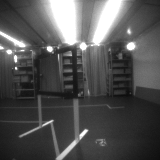}%
    \hfill
    \includegraphics[width=0.165\linewidth]{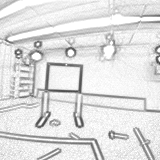}%
    \hfill
    \includegraphics[width=0.165\linewidth]{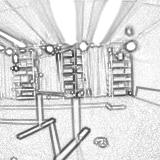}%
    \caption{Examples of (left) simulated images from \datastrain, (middle) real-world robot images in \datartrain, and (right) after applying the pencil filter~\cite{pham2022pencilnet}.}
    \label{fig:domain_examples}
\end{figure*}

We target the Crazyflie 2.1 Brushless by Bitcraze, a commercial off-the-shelf nano-drone extended with the pluggable GAP9Shield~\cite{gap9shield} and the Flow-deck expansion modules, depicted in Figure~\ref{fig:cf}.  
The GAP9Shield features a GreenWaves Technologies GAP9 ultra-low power System-on-Chip (SoC), composed of an eight-core RISC-V cluster (CL), an \SI{150}{MAC/cycle} 8-bit hardware accelerator for neural network inference (NE16), a cluster controller (CC) core in charge of orchestrating parallel computation, and a fabric controller core (FC) responsible for data movement with external peripherals.
GAP9 boasts a combined throughput up to \SI{220}{GOp/s} at max frequency \SI{370}{\mega\hertz}, while consuming less than \SI{100}{\milli\watt}.
The SoC integrates \SI{1.5}{\mega\byte} L2 memory and \SI{128}{\kilo\byte} L1 scratchpad, complemented on the GAP9Shield by off-chip \SI{32}{\mega\byte} OctaSPI RAM and \SI{64}{\mega\byte} Flash memories.
A 5MP OmniVision OV5647 camera provides visual perception, and the Espressif ESP32 module offers Wi-Fi connectivity.
The Crazyflie 2.1 base platform employs an STM32 microcontroller to handle low-level flight-control tasks and sensor interfacing:  
a 6-DoF IMU (three-axis accelerometer and gyroscope), an STMicroelectronics VL53L1X laser time-of-flight altitude sensor, and a PMW3901 optical-flow sensor to track horizontal displacement.  
The sensor data is fused by an Extended Kalman Filter~\cite{MuellerHamerUWB2015} to provide the onboard odometry.

\subsection{Data Collection}\label{sec:datacollection}

We collect training data in simulated and real-world environments.
In both domains, our $100\text{cm} \times 80\text{cm}$ gates are composed of four black-painted wooden beams placed at \SI{75}{\centi\meter} height,  based on the setup from the first Nanocopter AI challenge at IMAV 2022~\cite{lamberti2024droneracingsimtoreal}.
For the simulated data, we use the Webots simulator~\cite{Webots} to generate a dataset of 75K images depicting a gate in random poses.
To align the visual aspect of the simulated data in \datastrain\! with the images captured in the real world by the drone, we convert the images into grayscale and we perform data augmentations during training using Gaussian blur, vignetting, multiplicative Gaussian noise, and random exposure (see Figure~\ref{fig:domain_examples}).
%
%

To collect real-world images, we use the Crazyflie 2.1 Brushless piloted by a human to follow random trajectories around the gate while keeping it in the camera's FOV.
During flight, we collect images at 25 FPS with a resolution of $160 \times 160$ pixels. 
For testing purposes only, we collect ground truth pose data for both the drone and the gate with a motion capture system installed in our laboratory.
Real data is split into the training set \datartrain\! (51K samples), the validation set \datarval\! (8K samples), and the testing set \datartest\! (21K samples).
Examples of real-world images acquired by our drone are depicted in Figure~\ref{fig:domain_examples} (middle).

\subsection{Model Training and Onboard Deployment}\label{sec:model}
Our model architecture is a convolutional neural network with four convolutional blocks \notes{with 32, 32, 64, 128 channels, followed by batch normalization, and a linear layer with 128 neurons}, totaling 6M parameters.
For simulation training, we optimize the \lossp\! loss using AdamW~\cite{adamw} as the optimizer with a learning rate of $1e^{-3}$ for 100 epochs.
Then, we pick the model parameters with the lowest loss on \datasval\! and fine-tune it on real-world data by optimizing the \losssc\! loss using AdamW with a learning rate of $1e^{-6}$ for 100 epochs.
During fine-tuning, we use a small validation set \datarval, which does not contain pose labels, to pick the best model weights over the training process.

We deploy our model on the GWT GAP9 SoC to run entirely aboard our target nano-drone.
For maximum inference throughput, we aim to take advantage of the chip's NE16 neural accelerator, which only supports mixed-precision integer arithmetic.
We employ the GWT NNTool + AutoTiler deployment pipeline to quantize the model to 8-bit integers, with no loss in regression performance, and generate the C inference code.
This includes NE16 convolutional and fully-connected kernels, optimized parallel CPU kernels for the remaining layers, and automatically scheduled memory transfers across the chip's memory hierarchy.

\subsection{Evaluation Metrics}\label{sec:metrics}

We evaluate the model's ability to estimate the gate's relative pose on the real-world testing set \datartest by measuring the mean absolute error (MAE) and the Pearson correlation coefficient ($\rho$) between predicted and ground truth components of the 3D position and yaw of the gate; specifically for the yaw (indicated as $\psi$) angle, we consider the circular Pearson correlation coefficient~\cite{mardia2009directional}.
Due to the drone's flight dynamics, roll and pitch angles of the gate relative to the drone have a low variability in our data. As such, metrics on these variables would be uninformative; we choose not to present them in our evaluation.

Our model predictions correctly capture the pose of the gate but are inherently affected by a constant bias stemming from the nature of the state consistency loss:
The loss is minimized as long as the predicted pose pairs are spatially consistent with the drone's movements, yet they are not forced to be anchored at the gate's center.
This bias can be corrected with a calibration procedure by picking a small batch of images where the position of the gate relative to the camera is known.
The model is run on these images, and the mean offset between the gate's position and its estimate is measured.
To correct the bias, a constant offset is added to predicted positions.
In presenting our results, we apply this calibration procedure by considering the whole testing dataset.
This correction only affects the MAE metric for $x,y, \text{and}~z$, since the linear correlation coefficient is not affected by a constant bias.
For fairness, we apply the same bias correction procedure to all the considered baselines.

\subsection{Baselines}\label{sec:baselines}

\aboverulesep = 0mm 
\belowrulesep = 0mm %
\begin{table*}[!t]
    \centering
    \setlength\tabcolsep{1.2mm} 
    \renewcommand{\arraystretch}{1.2} 
    \caption{Mean absolute error MAE and Pearson correlation coefficient $\rho$ on the real-world testing set \textup{\datartest}, 3 runs per row.}
    \begin{tabular}{lcrrrrrrrrrr>{\centering\arraybackslash}p{40mm}}
    \toprule
    \multirow{2}{*}{Model} & \multirow{2}{*}{Supervision} & \multicolumn{1}{c}{MSE $\downarrow$} & \multicolumn{4}{c}{$\text{MAE} \downarrow$} & \multicolumn{4}{c}{$\rho \uparrow$} & {Plot for $\text{MSE}_{xyz}$ [cm] $\leftarrow$} \\
    \cmidrule(r){4-7}
    \cmidrule(l){8-11}
    & & $xyz$ [cm] & $x$ [cm] & $y$ [cm] & $z$ [cm] & $\psi$ [deg] & $x$ [\%] & $y$ [\%] & $z$ [\%] & $\psi$ [\%] & {Error bars mark 95\% CI} \\
    \specialrule{.4pt}{0.605mm}{0.984mm}
    \textit{Mean Predictor}                     & \datartest                    &  90.7 & 57.0 & 48.7 & 23.1 & 20.9 & $-$ & $-$ & $-$ & $-$ & \multirow{1}{*}{\input{tableplot.tikz}} \\
    \textit{Zero-Shot}                          & \datastrain                   &  79.7 & 51.6 & 41.0 & 24.5 & 33.5 & 38 & 53 & 40 &  5 & \\
    \textit{PencilNet~\cite{pham2022pencilnet}} & \datastrain                   &  88.3 & 56.7 & 46.8 & 22.5 & 32.7 &  4 & 40 & 44 &  2 & \\
    \textit{DA~\cite{fua2019sharingweights}}    & \datastrain{}, \datartrain{}  & 103.5 & 66.0 & 34.7 & 33.0 & 28.8 & 30 & 69 & 52 &  5 & \\
    \textit{Ours}                               & \datartrain{} (Odometry only) &  \textbf{44.7} & \textbf{25.6} & \textbf{28.2} & \textbf{10.5} & \textbf{13.1} & \textbf{88} & \textbf{79} & \textbf{91} & \textbf{82} & \\[1mm]
    \bottomrule
    \end{tabular}
    \label{tab:table}
\end{table*}

We validate our approach considering different baselines. 
The first one (\textit{Zero-Shot}) is our neural network trained only with the simulated dataset \datastrain{} and tested directly on \datartest. 
The second baseline we consider is \emph{PencilNet}~\cite{pham2022pencilnet}, a state-of-the-art domain generalization approach for gate pose estimation.
It consists of pre-processing the model's input images using a pencil filter to align the visual aspect of simulated and real-world data, as shown on the right of Figure~\ref{fig:domain_examples}. In this way, the model is trained in simulation and deployed to real scenarios in a zero-shot fashion. 
Our third baseline (\emph{DA}) is the well-established method for unsupervised domain adaptation described in~\cite{fua2019sharingweights}.
This approach uses a two-stream architecture to align the distributions of the source and target domains with Maximum Mean Discrepancy (MMD).
Furthermore, an additional loss function forces the parameters of the two streams to be linearly dependent. Similarly to our approach, no ground truth annotations in the real world are required.
To ensure a fair evaluation, we re-implement all the baselines using our model architecture (described in Section~\ref{sec:model}) and our training dataset acquired in simulation (see Section~\ref{sec:datacollection}) and report the performance figures obtained by averaging results over three training runs with different parameter initialization.
We also report the performance of the \textit{Mean Predictor}, which always predicts the average gate pose in the test set \datartest.

%
%
%
%
%
%
%

\section{Results}\label{sec:results}

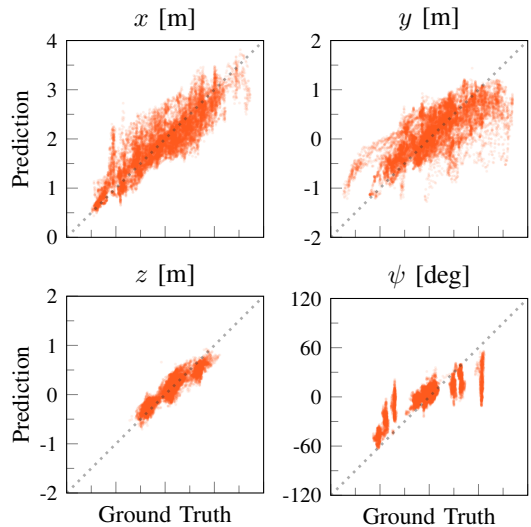
\begin{figure}[!t]%
    \centering%
    \begin{tikzpicture}
\def\scatterlen{42mm}
\begin{axis}[%
 name=scatter_x,
 axis equal,
 axis on top,
 height=\scatterlen,
 width=\scatterlen,
 title={$x$ [m]},
 ylabel={Prediction},
 title style={yshift=-2mm,},
 xmin=0.0, xmax=1.0,
 ymin=0.0, ymax=1.0,
 xtick={0.00, 0.25, 0.50, 0.75, 1.00},
 ytick={0.00, 0.25, 0.50, 0.75, 1.00},
 minor x tick num=1,
 minor y tick num=1,
 xticklabels=\empty,
 yticklabels={0, 1, 2, 3, 4},
 xtick pos=left,
 ytick pos=bottom,
 ticklabel style={font=\footnotesize},
 xlabel style={font=\small},
 ylabel style={font=\small},
]%
\addplot graphics[xmin=\pgfkeysvalueof{/pgfplots/xmin},ymin=\pgfkeysvalueof{/pgfplots/ymin},xmax=\pgfkeysvalueof{/pgfplots/xmax},ymax=\pgfkeysvalueof{/pgfplots/ymax}] {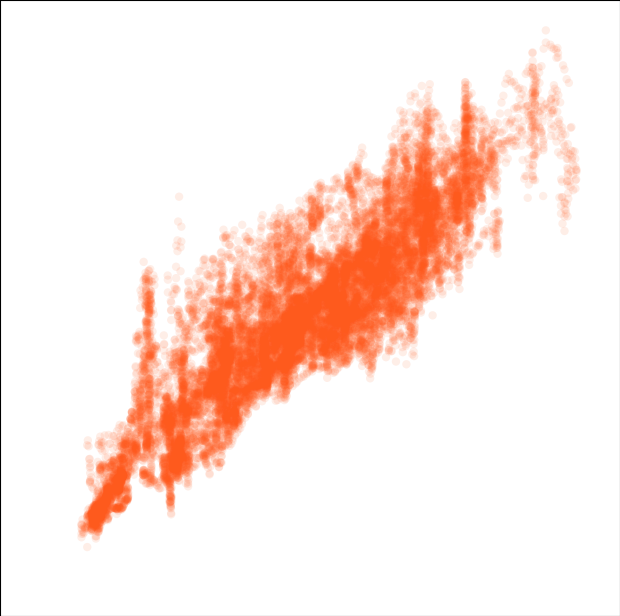};
\addplot[color=wong_black,line width=1.0pt,opacity=0.4,dotted] coordinates {(\pgfkeysvalueof{/pgfplots/xmin},\pgfkeysvalueof{/pgfplots/ymin}) (\pgfkeysvalueof{/pgfplots/xmax},\pgfkeysvalueof{/pgfplots/ymax})};%
\end{axis}%
\begin{axis}[%
 name=scatter_y,
 axis equal,
 axis on top,
 at=(scatter_x.right of north east),
 anchor=left of north west,
 xshift=3mm,
 height=\scatterlen,
 width=\scatterlen,
 title={$y$ [m]},
 title style={yshift=-2mm,},
 xmin=0.0, xmax=1.0,
 ymin=0.0, ymax=1.0,
 xtick={0.00, 0.25, 0.50, 0.75, 1.00},
 ytick={0.00, 0.25, 0.50, 0.75, 1.00},
 minor x tick num=1,
 minor y tick num=1,
 xticklabels=\empty,
 yticklabels={-2, -1, 0, 1, 2},
 xtick pos=left,
 ytick pos=bottom,
 ticklabel style={font=\footnotesize},
 xlabel style={font=\small},
 ylabel style={font=\small},
]%
\addplot graphics[xmin=\pgfkeysvalueof{/pgfplots/xmin},ymin=\pgfkeysvalueof{/pgfplots/ymin},xmax=\pgfkeysvalueof{/pgfplots/xmax},ymax=\pgfkeysvalueof{/pgfplots/ymax}] {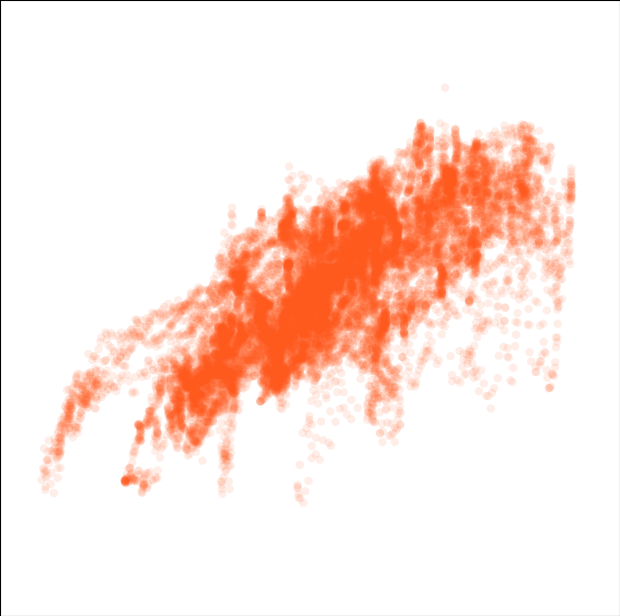};
\addplot[color=wong_black,line width=1.0pt,opacity=0.4,dotted] coordinates {(\pgfkeysvalueof{/pgfplots/xmin},\pgfkeysvalueof{/pgfplots/ymin}) (\pgfkeysvalueof{/pgfplots/xmax},\pgfkeysvalueof{/pgfplots/ymax})};%
\end{axis}%
\begin{axis}[%
 name=scatter_z,
 axis equal,
 axis on top,
 at=(scatter_x.below south west),
 anchor=above north west,
 height=\scatterlen,
 width=\scatterlen,
 title={$z$ [m]},
 title style={yshift=-2mm,},
 xlabel={Ground Truth},
 ylabel={Prediction},
 ylabel shift=-1mm,
 xmin=0.0, xmax=1.0,
 ymin=0.0, ymax=1.0,
 xtick={0.00, 0.25, 0.50, 0.75, 1.00},
 ytick={0.00, 0.25, 0.50, 0.75, 1.00},
 minor x tick num=1,
 minor y tick num=1,
 xticklabels=\empty,
 yticklabels={-2, -1, 0, 1, 2},
 xtick pos=left,
 ytick pos=bottom,
 ticklabel style={font=\footnotesize},
 xlabel style={font=\small, yshift=2mm,},
 ylabel style={font=\small},
]%
\addplot graphics[xmin=\pgfkeysvalueof{/pgfplots/xmin},ymin=\pgfkeysvalueof{/pgfplots/ymin},xmax=\pgfkeysvalueof{/pgfplots/xmax},ymax=\pgfkeysvalueof{/pgfplots/ymax}] {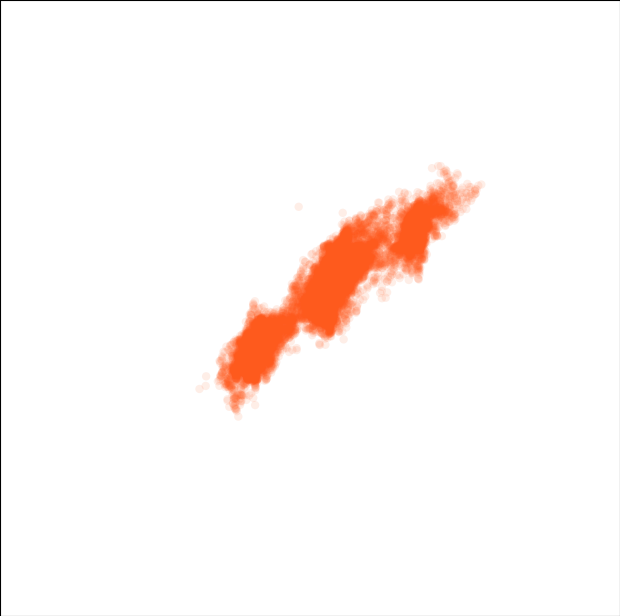};
\addplot[color=wong_black,line width=1.0pt,opacity=0.4,dotted] coordinates {(\pgfkeysvalueof{/pgfplots/xmin},\pgfkeysvalueof{/pgfplots/ymin}) (\pgfkeysvalueof{/pgfplots/xmax},\pgfkeysvalueof{/pgfplots/ymax})};%
\end{axis}%
\begin{axis}[%
 name=scatter_psi,
 axis equal,
 axis on top,
 at=(scatter_y.below south west),
 anchor=above north west,
 height=\scatterlen,
 width=\scatterlen,
 title={$\psi$ [deg]},
 title style={yshift=-2mm,},
 xlabel={Ground Truth},
 xmin=0.0, xmax=1.0,
 ymin=0.0, ymax=1.0,
 xtick={0.00, 0.25, 0.50, 0.75, 1.00},
 ytick={0.00, 0.25, 0.50, 0.75, 1.00},
 minor x tick num=1,
 minor y tick num=1,
 xticklabels=\empty,
 yticklabels={-120,-60,0,60,120},
 xtick pos=left,
 ytick pos=bottom,
 ticklabel style={font=\footnotesize},
 xlabel style={font=\small, yshift=2mm,},
 ylabel style={font=\small},
]%
\addplot graphics[xmin=\pgfkeysvalueof{/pgfplots/xmin},ymin=\pgfkeysvalueof{/pgfplots/ymin},xmax=\pgfkeysvalueof{/pgfplots/xmax},ymax=\pgfkeysvalueof{/pgfplots/ymax}] {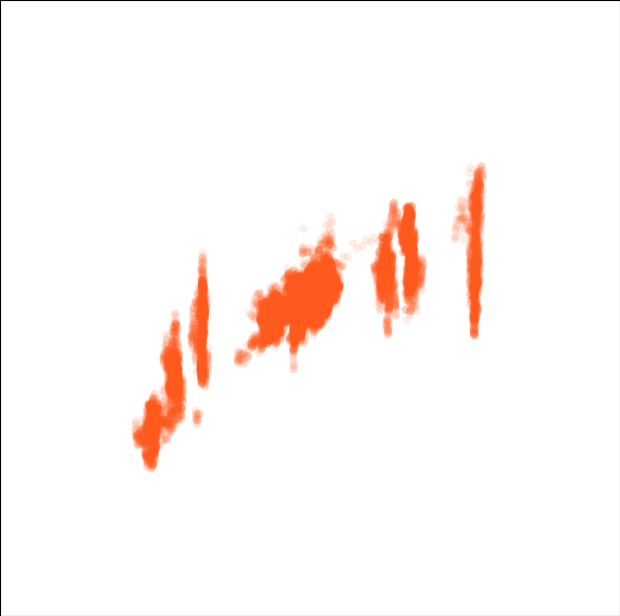};
\addplot[color=wong_black,line width=1.0pt,opacity=0.4,dotted] coordinates {(\pgfkeysvalueof{/pgfplots/xmin},\pgfkeysvalueof{/pgfplots/ymin}) (\pgfkeysvalueof{/pgfplots/xmax},\pgfkeysvalueof{/pgfplots/ymax})};%
\end{axis}
\end{tikzpicture}%
    \caption{\textit{Our} self-supervised domain adaptation approach vs ground truth on \datartest{} for the components of the gate pose ($x, y, z, \psi$).}%
    \label{fig:scatters}%
\end{figure}

We report the results for our approach and all baselines in Table~\ref{tab:table}.
%
We observe that our method consistently outperforms all the baselines:
The \textit{Zero-Shot} baseline, namely the model trained in simulation and tested in the real world, brings only marginal improvements with respect to the lower bound represented by the \emph{Mean Predictor}.
This is observed also with \emph{PencilNet} that, while reaching performance comparable to the \emph{Mean Predictor}, remarkably degrades the estimation of the gate's distance ($x$) in terms of MAE and $\rho$. This suggests that the domain generalization approach fails to overcome a large sim-to-real gap, such as the one found in our setup.
The results also show that our domain adaptation (\emph{DA}) baseline from~\cite{fua2019sharingweights} is not able to improve the performance of the \emph{Mean Predictor}.
Despite being beneficial for $y$, it degrades the estimation of both $x$ and $z$, demonstrating that performing distribution alignment is not sufficient to fill the domain gap of our scenario.
%
%
Compared with \emph{Mean Predictor}, our approach improves the MAE by $54\%$ in $x$ and $z$ (distance and height of the gate) and by $42\%$ in $y$ (horizontal displacement). Moreover, our approach is the only one that manages to reduce the error in the estimation of $\psi$, with a gain of $37\%$. 
Notably, the geometric relation between predictions imposed by the state consistency loss strongly improves the correlation $\rho$ of the model outputs with their respective ground truths (see Figure~\ref{fig:scatters}). 

\begin{figure*}[!t]%
    \centering%
    \includegraphics[width=0.2\linewidth]{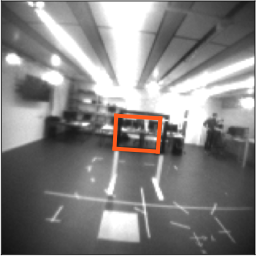}%
    \includegraphics[width=0.2\linewidth]{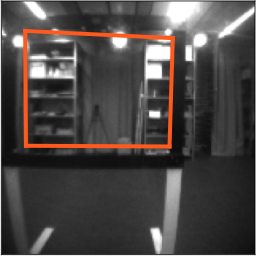}%
    \includegraphics[width=0.2\linewidth]{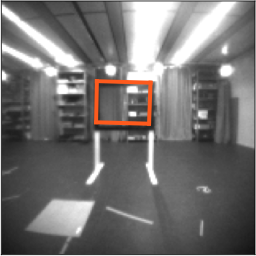}%
    \includegraphics[width=0.2\linewidth]{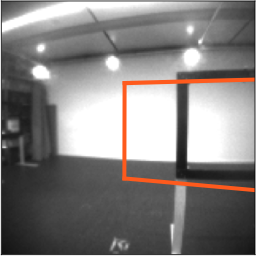}%
    \includegraphics[width=0.2\linewidth]{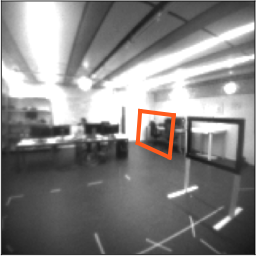}%
    \caption{Input frames overlayed with gate skeleton (in orange) using the pose predicted by our approach. The model correctly locates the gate (1-3), while failures occur when a large portion of the gate is occluded and for shallow angles (4-5).}%
    \label{fig:pose_regression_examples}%
\end{figure*}

Figure~\ref{fig:pose_regression_examples} reports qualitative examples of our model on gate pose estimation. In the first three images, the model precisely localizes the gate, even when it is far away from the drone. In the last two examples, challenging situations where the gate partially exits the image boundaries or is perceived from a narrow point of view cause the model to fail.

Following the pipeline described in Section~\ref{sec:model}, we deploy our network in the Crazyflie 2.1 quadrotor. Counting \SI{6}{\mega\nothing} parameters, the model requires \SI{318}{\mega\mac} per inference and \SI{30.4}{\milli\second} for processing a single frame on the NE16. With an inference throughput of \SI{32.8}{frames/\second}, our architecture is fully suitable for closed-loop autonomous control aboard the drone during real-world flights.

\subsection{Fine-tuning Data Ablation}

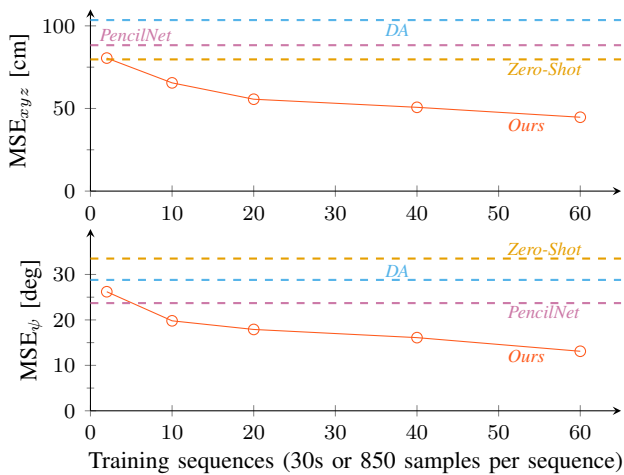
\begin{figure}[thbp]%
    \centering%
    \begin{tikzpicture}
%
\begin{axis}[
    name=performance,
    height=40mm,
    width=\linewidth,
    title style={yshift=-2mm,},
    ylabel={$\text{MSE}_{xyz}$ [cm]},
    ylabel shift=-1mm,
    xmin=0, xmax=65,
    ymin=0, ymax=110,
    minor x tick num=0,
    minor y tick num=1,
    axis x line=bottom,
    axis y line=left,
    ticklabel style={font=\footnotesize, color=black},
    xlabel style={font=\small, yshift=0.8mm,},
    ylabel style={font=\small},
]
    \addplot [giants_orange,mark=o] coordinates { (2, 80.5) (10, 65.5) (20, 55.6) (40, 50.7) (60, 44.7) };

    \addplot [domain=0:65, thick, dashed, wong_gold] {79.7};
    \addplot [domain=0:65, thick, dashed, wong_pink] {88.3};
    \addplot [domain=0:65, thick, dashed, wong_cyan] {103.5};

\node at (axis cs:50,73.7) [font={\scriptsize},anchor=mid west] {\color{wong_gold}\textit{Zero-Shot}};%
\node at (axis cs:0,94.3) [font={\scriptsize},anchor=mid west] {\color{wong_pink}\textit{PencilNet}};%
\node at (axis cs:35,97.5) [font={\scriptsize},anchor=mid west] {\color{wong_cyan}\textit{DA}};%
\node at (axis cs:50,40) [font={\scriptsize},anchor=mid west] {\color{giants_orange}\textit{Ours}};%

\end{axis}
%
\begin{axis}[
    at=(performance.below south west),
    anchor=above north west,
    height=40mm,
    width=\linewidth,
    ylabel shift=-1mm,
    xmin=0, xmax=65,
    ymin=0, ymax=40,
    axis y line=left,
    axis x line=bottom,
    minor x tick num=0,
    minor y tick num=1,
    xlabel={Training sequences (30s or 850 samples per sequence)},
    ylabel={$\text{MSE}_\psi$ [deg]},
    ticklabel style={font=\footnotesize, color=black},
    xlabel style={font=\small, yshift=0.8mm,},
    ylabel style={font=\small},
    ytick={0, 10, 20, 30},
]
    \addplot [giants_orange,mark=o] coordinates { (2, 26.2) (10, 19.8) (20, 17.9) (40, 16.1) (60, 13.1) };

    \addplot [domain=0:65, thick, dashed, wong_gold] {33.5};
    \addplot [domain=0:65, thick, dashed, wong_pink] {23.7};
    \addplot [domain=0:65, thick, dashed, wong_cyan] {28.8};

    \node at (axis cs:50,35.5) [font={\scriptsize},anchor=mid west] {\color{wong_gold}\textit{Zero-Shot}};%
    \node at (axis cs:50,21.7) [font={\scriptsize},anchor=mid west] {\color{wong_pink}\textit{PencilNet}};%
    \node at (axis cs:35,30.8) [font={\scriptsize},anchor=mid west] {\color{wong_cyan}\textit{DA}};%
    \node at (axis cs:50,12) [font={\scriptsize},anchor=mid west] {\color{giants_orange}\textit{Ours}};%
\end{axis}
\end{tikzpicture}%
    \caption{Position (top) and orientation (bottom) performance measure with MSE of our approach with increasing amounts of fine-tune data. DA is trained on the full real-world dataset (60 sequences), while PencilNet and Zero-Shot are trained only on \datastrain.}%
    \label{fig:tuning}%
\end{figure}

As a final experiment, we analyze the performance of our approach with varying amounts of target domain data: 1, 10, 20, 40, and 60 recording sequences, each one lasting 30 seconds and amounting to 850 samples.
Results reported in Figure~\ref{fig:tuning}, show that training our approach with 10 sequences is enough to surpass all of the baselines' performance.
By training with 40 sequences, we closely match the performance of the full dataset, showing diminishing returns, especially in estimating the gate's position.
These results prove that the approach is effective, and even small amounts of target domain data suffice to outperform state-of-the-art baselines.

\section{Conclusions}\label{sec:conclusion}
This work presents a self-supervised approach for gate pose estimation specifically tailored to mobile robots. Using solely the estimated odometry as the supervision signal for adapting a source-domain model, our method forces the model's predictions to be consistent with the drone's movements with the state consistency loss. The extensive experimental evaluation in the real world proves that our method outperforms both zero-shot and unsupervised domain adaptation baselines. In addition, we demonstrate that our approach is effective, even when a limited amount of target domain data is available.
Lastly, \notes{we assess the effectiveness of our approach on a lightweight neural network that runs at 33 fps onboard the Crazyflie 2.1 nano-drone.}
In future work, we plan to extend our approach with few-shot adaptation techniques to further reduce the amount of target domain data, and introduce uncertainty estimation into the training setup, aiming to minimize the entropy in a similar fashion to source-free domain adaptation~\cite{kim2021domain}.
\notes{Furthermore, we plan to validate our approach in a closed-loop drone racing setup.}

\addtolength{\textheight}{-2.3cm}   

\bibliographystyle{IEEEtran}
\bibliography{references}

\end{document}